\documentclass[lettersize,journal]{IEEEtran} 
\usepackage{amsmath,amsfonts}
\usepackage{algorithm}
\usepackage{algpseudocode}
\usepackage{array}
\usepackage[caption=false,font=normalsize,labelfont=sf,textfont=sf]{subfig}
\usepackage{textcomp}
\usepackage{stfloats}
\usepackage{url}
\usepackage{verbatim}
\usepackage{graphicx}
\usepackage{cite}
\usepackage{amsmath}
\usepackage{amssymb}
\usepackage{xcolor}
\usepackage{booktabs}
\usepackage{multirow}
\usepackage{bbding}
\usepackage{orcidlink}
\usepackage{bbm}
\usepackage{colortbl}
\usepackage[capitalise]{cleveref}
\begin{document}

\title{DUSE: A Data Expansion Framework for Low-resource Automatic Modulation Recognition based on Active Learning}

\author{Yao Lu\orcidlink{0000-0003-0655-7814}, ~\IEEEmembership{Student Member, IEEE}, Hongyu Gao, Zhuangzhi Chen, Dongwei Xu\orcidlink{0000-0003-2693-922X},~\IEEEmembership{Member, IEEE}, Yun Lin\orcidlink{0000-0003-1379-9301},~\IEEEmembership{Senior Member, IEEE}, Qi Xuan\orcidlink{0000-0002-6320-7012},~\IEEEmembership{Senior Member,~IEEE}, Guan Gui,~\IEEEmembership{Fellow,~IEEE}
\thanks{This work was partially supported by the National Natural Science Foundation of China under Grant 62301492, 61973273, U21B2001 and by the Key R\&D Program of Zhejiang under Grant 2022C01018. (Corresponding author: Qi Xuan, Dongwei Xu)}
\thanks{Yao Lu is with the Institute of Cyberspace Security, College of Information Engineering, Zhejiang University of Technology, Hangzhou 310023, China, with the Binjiang Institute of Artificial Intelligence, Zhejiang University of Technology, Hangzhou 310056, China, also with the Centre for Frontier AI Research, Agency for Science, Technology and Research, Singapore 138632 (e-mail: yaolu.zjut@gmail.com).}
\thanks{Hongyu Gao, Dongwei Xu and Qi Xuan are with the Institute of Cyberspace Security, College of Information Engineering, Zhejiang University of Technology, Hangzhou 310023, China, also with the Binjiang Institute of Artificial Intelligence, Zhejiang University of Technology, Hangzhou 310056, China (e-mail: ghybgyylj@gmail.com, dongweixu@zjut.edu.cn, xuanqi@zjut.edu.cn).}
\thanks{Zhuangzhi Chen is with the Institute of Cyberspace Security, Zhejiang University of Technology, Hangzhou 310023, China, and also with the Research Center of Electromagnetic SpaceSecurity, Binjiang Institute of Artificial Intelligence, ZJUT, Hangzhou, 310056, China (e-mail: zzch@zjut.edu.cn)}
\thanks{Yun Lin is with the College of Information and Communication Engineering, Harbin Engineering University, Harbin, China (e-mail: linyun@hrbeu.edu.cn).}
\thanks{Guan Gui is with the College of Telecommunications and Information Engineering, Nanjing University of Posts and Telecommunications, Nanjing 210003, China (e-mail: guiguan@njupt.edu.cn).}}

\markboth{Journal of \LaTeX\ Class Files,~Vol.~14, No.~8, August~2021}%
{Shell \MakeLowercase{\textit{et al.}}: A Sample Article Using IEEEtran.cls for IEEE Journals}


\maketitle

\begin{abstract}
Although deep neural networks have made remarkable achievements in the field of automatic modulation recognition (AMR), these models often require a large amount of labeled data for training. However, in many practical scenarios, the available target domain data is scarce and difficult to meet the needs of model training. The most direct way is to collect data manually and perform expert annotation, but the high time and labor costs are unbearable. Another common method is data augmentation. Although it can enrich training samples to a certain extent, it does not introduce new data and therefore cannot fundamentally solve the problem of data scarcity. To address these challenges, we introduce a data expansion framework called \textbf{D}ynamic \textbf{U}ncertainty–driven \textbf{S}ample \textbf{E}xpansion (\textbf{DUSE}). Specifically, DUSE uses an uncertainty scoring function to filter out useful samples from relevant AMR datasets and employs an active learning strategy to continuously refine the scorer. Extensive experiments demonstrate that DUSE consistently outperforms $8$ coreset selection baselines in both class‐balance and class‐imbalance settings. Besides, DUSE exhibits strong cross‑architecture generalization for unseen models.
\end{abstract}

\begin{IEEEkeywords}
Automatic Modulation Recognition, Data Expansion, Coreset Selection, Active Learning, Generalization
\end{IEEEkeywords}

\section{Introduction}
\IEEEPARstart{I}{n} recent years, the rapid development of deep learning~\cite{wu2023towards,huang2022feature,liu2024fine} has significantly advanced the field of Automatic Modulation Recognition (AMR), transforming it from methods~\cite{walenczykowska2016type,li2019wavelet,triantafyllakis2017phasma,vuvcic2017cyclic,abdelmutalab2016automatic,yao2025minimizing} relying on feature engineering to an end-to-end learning framework~\cite{o2018over,o2016convolutional,lin2020improved,chen2021signet,lin2020contour,tu2020complex,lin2020adversarial,zhang2023lightweight,hou2024multi,xu2025mclrl,lu2025fcos,lu2024generic} that can interpret complex signal characteristics. For example, inspired by the VGG~\cite{simonyan2014very} architecture in image classification, O'Shea et al.~\cite{o2018over} introduce a 1D-CNN model for short radio signal classification, and later extend it to a lightweight 2D‐CNN processing both in‐phase (I) and quadrature (Q) components in the time domain~\cite{o2016convolutional}. Chen et al.~\cite{chen2021signet} further enhance feature representations via the S2M operator, which converts raw signals into square matrices to enable the use of advanced image‐based classification models. To tackle low signal‐to‐noise ratio (SNR) scenarios, Zhang et al.~\cite{zhang2023amc} propose AMC-Net, which denoises in the frequency domain and performs multi-scale feature extraction for robust classification. Despite these advances, training such deep learning models requires large annotated datasets, yet in many AMR applications only limited target‐domain samples are available, motivating the need to expand the effective training data.

The most straightforward solution is to manually collect and label signal samples. However, this is often impractical in real-world environments, as it requires expensive equipment~\cite{hanna2022wisig,reus2020trust} (e.g. USRP and ADS-B) and professional engineers to analyze each sample, consuming a lot of time and manpower. A common alternative is to apply data augmentation techniques~\cite{shen2024simple,wei2023efficient,patel2020data,zheng2021spectrum,chen2024data,li2025diffusion} such as time shifting, noise injection, or mixing, but these schemes only create synthetic variants of existing samples without introducing real signal samples, thus failing to address the fundamental scarcity problem. In contrast, augmenting with auxiliary samples drawn from large‐scale AMR datasets (e.g., RML2018.01a~\cite{o2018over}) can substantially enlarge the training dataset with realistic signal examples and mitigate overfitting than classical augmentation.

Inspired by this, we propose a data expansion framework, named Dynamic Uncertainty–driven Sample Expansion (DUSE), which selectively transfers samples from readily available auxiliary datasets of the relevant AMR domain to the target dataset. Specifically, DUSE introduces an uncertainty scoring function to quantify the potential contribution of each sample in the auxiliary dataset and employs an active learning strategy to iteratively update the model for scoring and refine its uncertainty estimate to adapt to the changing data distribution. By selecting a small relevant subset of the auxiliary dataset and merging it into the target training set, DUSE not only increases the diversity and quantity of training data, but also preserves the distribution characteristics of the target dataset.

To evaluate the effectiveness of DUSE, we conduct comprehensive experiments across three widely-used benchmark datasets (RML2016.10a~\cite{o2016radio}, Sig2019-12~\cite{chen2021signet}, and RML2018.01a~\cite{o2018over}) against $8$ coreset selection baselines. Extensive experiments show that DUSE outperforms existing baselines in both class-balanced and class-imbalanced settings and exhibits good cross-architecture generalization. Experiments on the t-SNE visualization of the expanded dataset and statistics of the number of samples in each category further verify the reasons why our method can achieve good results.

To summarize, our main contributions are three-fold:
\begin{itemize}
\item[$\bullet$] We first propose the concept of data expansion in the AMR field and give a formal definition.
\item[$\bullet$] We introduce a novel data expansion framework DUSE, which uses an uncertainty scoring function to quantify the potential contribution of each sample in the auxiliary dataset and employs an active learning strategy to iteratively update the model for scoring.
\item[$\bullet$] Extensive experiments on $3$ AMR datasets show that DUSE outperforms $8$ coreset selection baselines in both class-balanced and class-imbalanced settings and exhibits good cross-architecture generalization capabilities.
\end{itemize}

In the remainder of this paper, we first introduce related works on coreset selection and active learning in \cref{sec:Related Works}. In \cref{sec:model purify}, we delve into our Dynamic Uncertainty–driven Sample Expansion (DUSE) framework. Then experiments are discussed in \cref{sec:Experiments}. Finally, the paper concludes in \cref{sec:Conclusion}.

\section{Related Works}
\label{sec:Related Works}
\textbf{Coreset Selection.} Coreset selection, also known as data pruning, seeks to extract a compact yet highly informative subset of training examples that maintains the original model’s performance while cutting down on computational expense. In practice, most methods first define a scoring function tailored to the task and data characteristics, then rank every example in the dataset according to that score. Finally, they pick the top‑ranked points (or those within a specified score range) to form the coreset for all subsequent training or evaluation. For example, early coreset selection methods are designed for traditional machine learning algorithms, such as K-means~\cite{har2005smaller,feldman2011unified}, SVM~\cite{tsang2005core}, logistic regression~\cite{huggins2016coresets}, and Gaussian mixture model~\cite{lucic2018training,feldman2011scalable}. However, these techniques are difficult to directly apply to deep neural networks. After that, Chai et al.~\cite{chai2023efficient} cluster the training set via euclidean distance, bound the full gradient using the maximum feature distance between each item and each cluster, and perform selection by iterating through these clusters. Guo et al.~\cite{guo2022deepcore} and Moser et al.~\cite{moser2025coreset} provide an empirical study on multiple coreset selection methods. Lu et al.~\cite{lu2024rk} introduce RK-core to empower gaining a deeper understanding of the intricate hierarchical structure within datasets. Recently, Lee et al.~\cite{lee2024coreset} pioneer the application of coreset selection to object detection tasks. Meanwhile, Tong et al.~\cite{tong2025coreset} and Hao et al.~\cite{hao2025fedcs} establish coreset selection frameworks for continuous learning scenarios, while Hao et al.~\cite{hao2025fedcs} explore coreset selection in federated learning scenarios. Chai et al.~\cite{chai2023goodcore} investigate the problem of coreset selection over incomplete data.

Despite significant progress in coreset selection for computer‑vision tasks, its application to AMR remains largely unexplored. This gap stems from fundamental differences between signal and image data, which render image‑oriented coreset selection methods ineffective or outright inapplicable on signal datasets.

\textbf{Active Learning.} Active learning is a long-studied paradigm aimed to maximize a model’s performance gain while annotating the fewest samples possible~\cite{settles2009active,yang2024learning}. For example, Zhang et al.~\cite{zhang2022cost} propose a SVR-based active learning framework tailored for time series forecasting, with the objective of selecting the most informative samples while minimizing training set complexity. Peng et al.~\cite{peng} incorporates cross-entropy loss into a bidirectional gated recurrent neural network framework, forming a cost-sensitive active learning approach tailored for imbalanced fault diagnosis. Kim et al.~\cite{kim2021task} propose task-aware variational adversarial active learning, which considers data distribution of both label and unlabeled pools, relaxes task learning loss prediction to ranking loss prediction and uses ranking conditional generative adversarial network to embed normalized ranking loss information on VAAL. Deng et al.~\cite{deng2018active} propose a unified deep network, combined with active transfer learning that can be well-trained for hyperspectral images classification using only minimally labeled training data. Besides, Bhargava et al.~\cite{bhargava2022modulation} extend active learning to the modulation classification task and use an active learning framework to solve the real-time wireless modulation and signal class labeling problem. 

Among these methods, the most relevant to ours is the work by Bhargava et al.~\cite{bhargava2022modulation}. However, unlike their setting, which focuses on labeling unlabeled data through an active learning framework, our problem formulation is fundamentally different: we aim to select a subset from an already labeled auxiliary dataset to augment the target training set under a fixed sampling budget.

\subsection{Problem Definition}
In many practical tasks, researchers usually have access only to a small, truly annotated target‐domain dataset $\mathcal{D}_T \in \{x_i^T, y_i^T\}_{i=1}^{N_T}$, where $N_T$ is very small (e.g., hundreds to thousands samples), making it insufficient to train deep learning models from scratch. To overcome the problems of overfitting and unstable training caused by the scarcity of samples, we hope to use a ready-made large-scale auxiliary dataset $\mathcal{D}_A$ to augment $\mathcal{D}_T$, where
\begin{equation}
  \mathcal{D}_A\in \{x_i^A, y_i^A\}_{i=1}^{N_A}, \quad N_A \gg N_T.
\end{equation}
This auxiliary dataset comes from a related but different field (e.g., general automatic modulation recognition datasets~\cite{o2016radio,o2018over,chen2021signet}). By picking the samples most relevant to the target domain and merging them into $\mathcal{D}_T$, the amount of training samples can be greatly increased while maintaining the target distribution characteristics, thereby improving the generalization performance of the model. 

In summary, we can formalize the above into the following problem: given a budget $K$ (the maximum number of auxiliary samples to select), we aim to choose a subset \(\mathcal{S}^* \subset \mathcal{D}_A\) such that the augmented training set
\begin{equation}
  \mathcal{D}_{\mathrm{aug}} = \mathcal{D}_T \cup \mathcal{S}^*, \quad |\mathcal{S}^*| \leq K,
\end{equation}
maximizes classification accuracy on the target dataset:
\begin{equation}
   \mathop{min}\limits_{\theta} \frac{1}{|\mathcal{D}_\mathrm{aug}|} \sum_{i=1}^{|\mathcal{D}_\mathrm{aug}|}{\mathcal{L}\bigl(f(x_i; \theta), y_j\bigr)}, \quad 
  \label{eq:classification}
\end{equation} 
where $f(\cdot;\theta)$ denotes the deep learning models parameterized by $\theta$ and $\mathcal{L}(\cdot)$ denotes the loss function.

\begin{figure*}[t]
	\centering
    \includegraphics[width=0.99\textwidth]{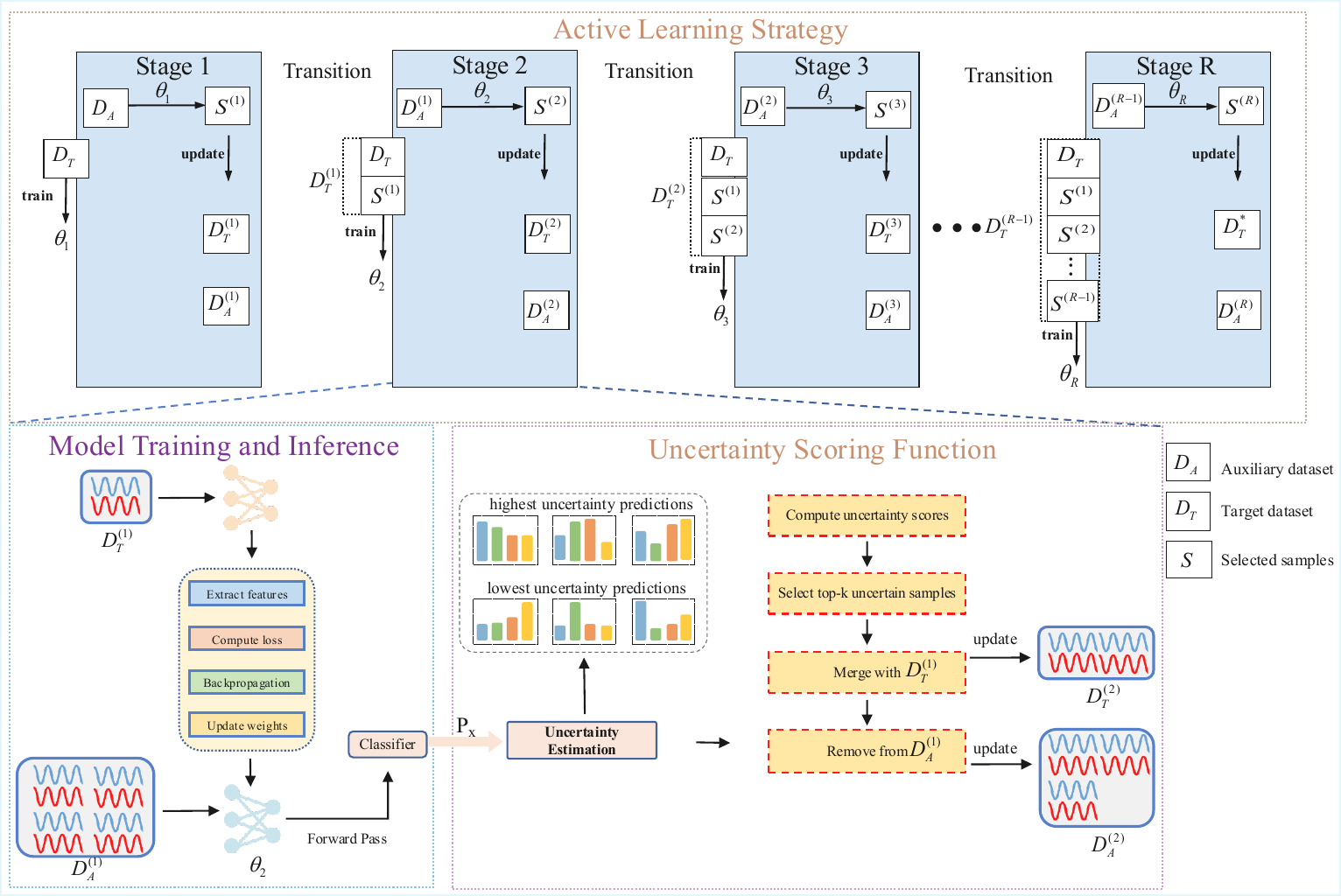} 
 \caption{\textbf{Framework Overview}. DUSE is consist of two components: (1) an \textbf{Uncertainty Scoring Function} that estimates the informativeness of each sample relative to the target dataset, and (2) an \textbf{Active Learning Strategy} that progressively augments the target dataset using this score.} 
 \label{fig:DUSE}
\end{figure*}

\begin{algorithm}[t]
\caption{Uncertainty Scoring Function}\label{alg:uncertainty_scoring}
\textbf{Input}: A deep learning model $f(\cdot; \theta)$, input sample $x$, number of classes $C$.\\
\textbf{Output}: Uncertainty score $u$. \\
\begin{algorithmic}[1]
\State $F = f(x; \theta)$ \Comment{Forward pass to get logits $F \in \mathbb{R}^C$}
\State $P \gets [0]^C$ \Comment{Initialize probability vector}
\For{$j = 1$ \textbf{to} $C$}
    \State $p_j \gets \exp(F_j) / \sum_{c=1}^C \exp(F_c)$ \Comment{Compute softmax}
\EndFor
\State $P_{\downarrow} = \textsc{Sort}(P, \text{descending})$ \Comment{$[p_1^* \geq p_2^* \geq \cdots \geq p_C^*]$}
\State $u \gets p_1^* - p_2^*$ \Comment{Uncertainty score (Eq. \ref{eq:margin_score})}
\end{algorithmic}
\end{algorithm}

\section{Method}
\label{sec:model purify}
To address the challenge of limited labeled data in target domains, we propose a Dynamic Uncertainty–driven Sample Expansion (DUSE) framework based on active learning. DUSE iteratively expands a small, high-quality labeled dataset ($\mathcal{D}_T$) by selectively incorporating the most informative samples from an auxiliary dataset $\mathcal{D}_A$. Specifically, DUSE is consist of two components: (1) an \textbf{Uncertainty Scoring Function} that estimates the informativeness of each sample relative to the target dataset, and (2) an \textbf{Active Learning Strategy} that progressively augments the target dataset using this score. We provide the overall process of DUSE in \cref{fig:DUSE}. Next, we will elaborate on these two core components.

\begin{algorithm}[t]
    \caption{Dynamic Uncertainty–driven Sample Expansion}
    \label{algorithm:Dynamic Uncertainty–driven Sample Expansion}
    \textbf{Input}: A target dataset $\mathcal{D}_T \in \{x_i^T, y_i^T\}_{i=1}^{N_T}$, an auxiliary dataset $\mathcal{D}_A\in \{x_i^A, y_i^A\}_{i=1}^{N_A}$, training epoch $E$, iterative round $R$, a deep learning model $f(\cdot; \theta)$.\\
    \textbf{Output}: An expanded dataset $\mathcal{D}_{T}^*$. \\
    \begin{algorithmic}[1] 
    \For{$i = 1$ to $R$}
        \State Train model $f(\cdot; \theta)$ on $\mathcal{D}_T$ for $E$ epochs using \cref{eq:celoss}. 
        \For{$j = 1$ to $N_A$ in $\mathcal{D}_A$}
        \State Compute uncertainty score $u_j$ for $x_j^A$ via Eq.~\eqref{eq:margin_score}
        \EndFor
        \State Obtain $\mathbf{u}^{(1)}=\left[u_1, u_2, \ldots, u_{N_A}\right]^{\top}$
        \State Sort $\mathbf{u}^{(1)}$ in descending order
        \State Identify the top-$K$ most uncertain samples $\mathcal{S}^{(1)}$
        \State Migrate $\mathcal{S}^{(1)}$ to the target dataset $\mathcal{D}_{T}$ 
        \State Update the auxiliary dataset $\mathcal{D}_A$ to remove $\mathcal{S}^{(1)}$:
    \EndFor
    \State Obtain $\mathcal{D}_{T}^* \leftarrow \mathcal{D}_{T} \cup \mathcal{S}^{(1)} \cup \cdots \cup \mathcal{S}^{(R)}$
    \end{algorithmic}
\end{algorithm}

\subsection{Uncertainty Scoring Function}
\label{sec:uncertainty}
\begin{figure}[t]
	\centering
    \includegraphics[width=0.49\textwidth]{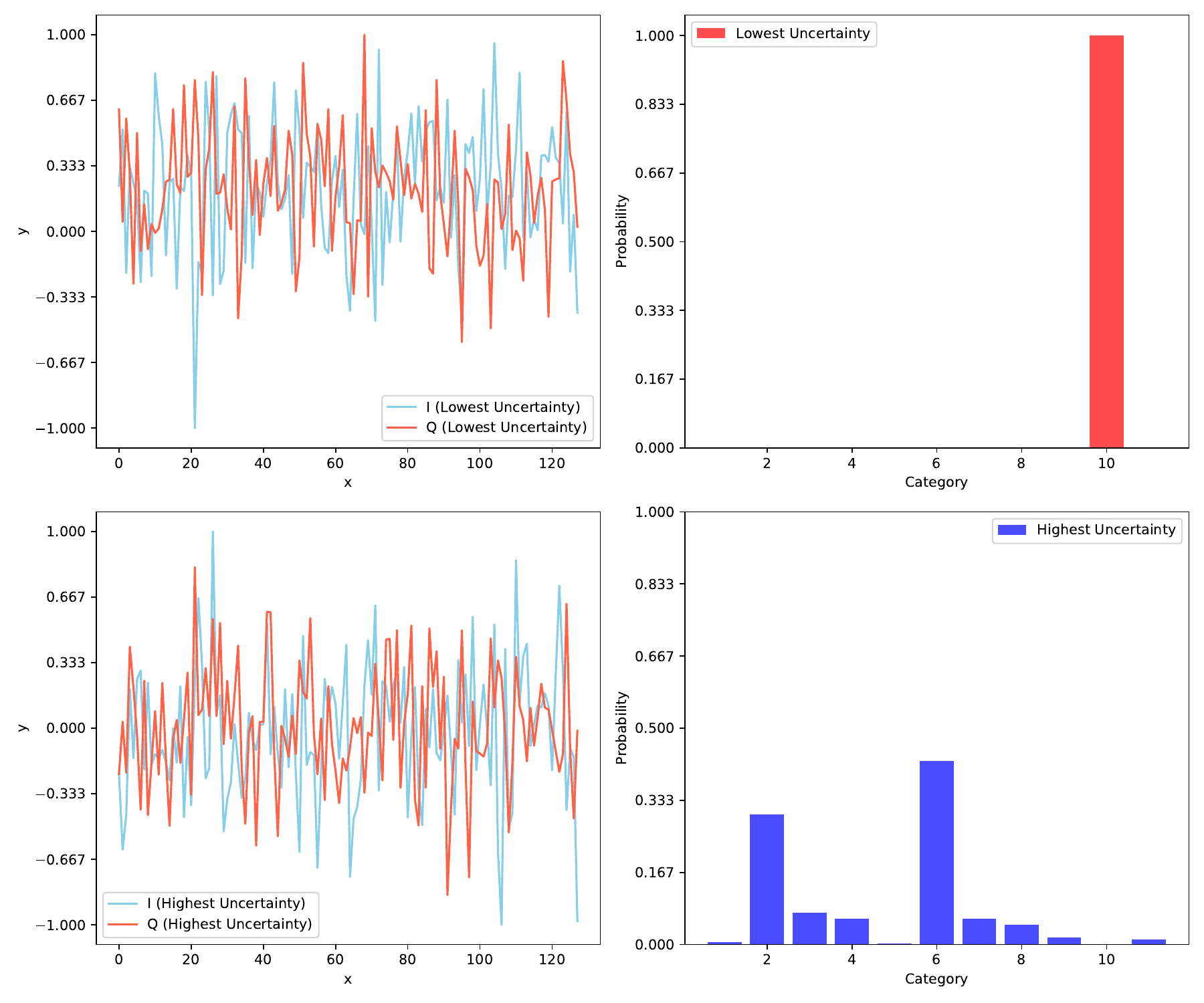} 
 \caption{Visualization of the samples with the highest and lowest uncertainty and their class-probability vectors.} 
 \label{fig:example}
\end{figure}

To quantify how “informative” each sample in $\mathcal{D}_A$ is, we introduce a \emph{Uncertainty Scoring Function}. Specifically, let $f(\cdot;\theta)$ be a deep learning model parameterized by $\theta$, we first feed an example $x \in \mathcal{D}_A$ into the model to obtain its logits:
\begin{equation}
    F = f(x;\theta) \in \mathbb{R}^C,
\end{equation}
where $C$ denotes the number of classes. We then convert the logits to a class‐probabilities vector using the softmax function:
\begin{equation}
    P = \left[ p_1, p_2, \dots, p_C \right]^\top, \quad \text{where} \quad p_j = \frac{\exp(F_j)}{\sum_{c=1}^C \exp(F_c)}.
    \label{eq:softmax}
\end{equation}
To quantify uncertainty, we sort its class probabilities $P = [p_{1}, p_{2}, \dots, p_{C}]^\top$ in descending order to obtain:
\begin{equation}
P_{\downarrow} = [p_{1}^{*}, p_{2}^{*}, \dots, p_{C}^{*}]^\top,
\quad \text{where} \quad
p_{1}^{*} \geq p_{2}^{*} \geq \cdots \geq p_{C}^{*}.
\end{equation}
The uncertainty scoring function for $x$ is then defined as:
\begin{equation}
  u = p_{1}^{*} - p_{2}^{*}.
  \label{eq:margin_score}
\end{equation}
A smaller $u$ indicates higher model uncertainty on $x$. As shown in \cref{fig:example}, to better understand the uncertainty value, we visualize the samples with the highest and lowest uncertainty values in the same category and their corresponding probability vectors. The reason why we choose $u$ as our data selection metric is that samples with small $u$ lie near decision boundaries and the decision boundaries can be better refined using these samples. Finally, we provide the detail description of the process in \cref{alg:uncertainty_scoring}.

\subsection{Dynamic Uncertainty–driven Sample Expansion}
\label{sec:duse}
After obtaining the uncertainty scoring function to select the most informative samples, we implement a Dynamic Uncertainty–driven Sample Expansion framework based on the active learning strategy. This framework progressively transfers low-value samples from the auxiliary dataset $\mathcal{D}_A$ (rest set) to the target training set $\mathcal{D}_T$ through iterative refinement. 

Specifically, given an initial deep learning model $f(\cdot;\theta)$, we first train it using the target dataset $\mathcal{D}_{T}$ via empirical risk minimization with $E$ training epochs:
\begin{equation}
\arg \min _\theta \sum_{(x, y) \in \mathcal{D}_{T}} \mathcal{L}(f(x; \theta), y),
\label{eq:celoss}
\end{equation}
where $\mathcal{L}$ denotes the cross-entropy loss function in this paper. Then we compute the uncertainty scores using \Cref{eq:margin_score}) for all samples in the auxiliary dataset $\mathcal{D}_A$:
\begin{equation}
\mathbf{u}^{(1)}=\left[u_1, u_2, \ldots, u_{N_A}\right]^{\top}
\end{equation}
Subsequently, we sort $\mathbf{u}^{(1)}$ in descending order and identify the top-$K$ most uncertain samples $\mathcal{S}^{(1)}$:
\begin{equation}
\mathcal{S}^{(1)}=\left\{\left(x_j, y_j\right) \mid u_j \in \text {top-} K \min \left(\mathbf{u}^{(1)}\right)\right\}.
\end{equation}
We migrate the selected samples $\mathcal{S}^{(1)}$ to the target dataset $\mathcal{D}_{T}$ and update the auxiliary dataset $\mathcal{D}_A$ to remove these samples:
\begin{equation}
\left\{\begin{array}{l}
\mathcal{D}_{T}^{(1)} \leftarrow \mathcal{D}_{T} \cup \mathcal{S}^{(1)} \\
\mathcal{D}_{A}^{(1)} \leftarrow \mathcal{D}_A \backslash \mathcal{S}^{(1)}.
\end{array}\right.
\end{equation}
The deep learning model is then retrained on the updated target dataset $\mathcal{D}_{T}^{(1)}$. We repeat the above operation for $R$ rounds and finally obtain the updated target dataset $\mathcal{D}_{T}^*$:
\begin{equation}
\mathcal{D}_{T}^* \leftarrow \mathcal{D}_{T} \cup \mathcal{S}^{(1)} \cup \cdots \cup \mathcal{S}^{(R)}.
\end{equation}
Finally, we provide the detail description of our DUSE framework in \cref{algorithm:Dynamic Uncertainty–driven Sample Expansion}.

\section{Experiments}
\label{sec:Experiments}
In this section, we verify the effectiveness of DUSE against $6$ baseline methods on $6$ deep learning-based AMR models across $3$ widely-used AMR datasets. Next, we will introduce the datasets and baselines used in detail.

\subsection{Datasets}
In this paper, we conduct extensive experiments on three AMR benchmarks: RML2016.10a~\cite{o2016radio}, Sig2019-12~\cite{chen2021signet} and RML2018.01a~\cite{o2018over}. Below is an overview of each dataset:

\textbf{RML2016.10a.} This dataset is generated by GNU Radio~\cite{blossom2004gnu} and contains $11$ modulation types and a total of $220,000$ samples. Each signal-to-noise ratio (SNR) for each modulation type contains $1,000$ samples of length $128$. The SNR varies from $-20$dB to $+18$dB in $2$dB steps. We partition the dataset into training and test splits in a $8:2$ ratio for model evaluation.

\textbf{Sig2019-12}~\cite{chen2021signet} contains $468,000$ simulated signals across $12$ modulation types, each of length $512$. Its SNR varies from $-20$dB to $+30$dB at $2$dB intervals, with $1,500$ samples per SNR for each modulation type. We partition the dataset into training and test splits in a $2:1$ ratio.

\textbf{RML2018.01a}~\cite{o2018over} contains more than $2.55$ million samples covering $24$ modulation types, which are recorded under laboratory conditions using simulated wireless channels. The signal length is $1,024$ points, and the SNR ranges from $-20$dB to $+30$dB with a step size of $2$dB. RML2018.01a applies the same data partitioning ratio as RML2016.10a.

In this paper, we only use samples with SNR$>10$dB to conduct experiments. 

\textbf{Composition of the auxiliary dataset and target dataset.} The target dataset $\mathcal{D}_T$ is initialized as a $1\%$ class-balanced random subset of the full training dataset. The auxiliary dataset $\mathcal{D}_A$ constitutes the remaining $99\%$ of training samples, maintaining identical per-class proportions to ensure distributional equivalence.

\subsection{Baselines}
In this paper, we compare our DUSE with $8$ existing state-of-the-art coreset selection approaches. We provide a detailed description of these methods as follows:

\textbf{Least confidence.} Measure uncertainty as $1-\text{max } P(\hat{y}|x)$, where higher values indicate lower confidence in the top predicted class.

\textbf{Entropy.} Quantify prediction uncertainty via Shannon entropy (\cref{eq:Entropy}), where higher values reflect more uniform probability distributions.
\begin{equation}
u_{\text{entropy}}(x)=-\sum_{i=1}^C P(\hat{y}=i \mid x) \log P(\hat{y}=i \mid x)
\label{eq:Entropy}
\end{equation}

\textbf{Margin~\cite{coleman2019selection}.} Compute $1 - (P(\hat{y}|x) - \max_{y \neq \hat{y}} P(y|x))$, with smaller margins (higher scores) indicating greater confusion between top two classes.

\textbf{DeepFool.} Since exact distance to the decision boundary is inaccessible, Ducoffe et al.~\cite{ducoffe2018adversarial} try to approximate these distances in the input space. By perturbing samples until the predictive labels of samples are changed, those data points require the smallest adversarial perturbation are closest to the decision boundary.

\textbf{Forgetting.} Toneva et al.~\cite{toneva2018empirical} count how many times the forgetting
happens during the training, i.e. the misclassification of a sample in the current epoch after having been correctly classified in the previous epoch, formally $acc^e_i > acc^e_{i+1}$, where $acc^e$ indicates the correctness (True or False) of the prediction of sample $i$ at epoch $e$. The number of forgetting reveals intrinsic properties of the training data, allowing for the removal of unforgettable examples with minimal performance drop.

\textbf{Glister.} Coreset selection can be treated as a bilevel optimization problem, therefore Glister et al~\cite{killamsetty2021glister} introduce a validation set $\mathcal{V}$ on the outer optimization:
\begin{equation}
\mathcal{S}^*=\underset{\mathcal{S} \subset \mathcal{D}_A}{\arg \max } \sum_{(x, y) \in \mathcal{V}} \mathcal{L}\left(x, y ; \underset{\theta}{\arg \max } \sum_{(x, y) \in \mathcal{S}} \mathcal{L}(x, y ; \theta)\right).
\end{equation}

\textbf{GraNd.} The GraNd score~\cite{paul2021deep} of sample $(x, y)$ at epoch $e$ is defined as:
\begin{equation}
u_{\text{GraNd}}(x)\triangleq \mathbb{E}_{\theta_e}\left\|\nabla_{\theta_e} \mathcal{L}\left(x, y;\theta_e\right)\right\|_2 .
\end{equation}
It measures the average contribution from each sample to the decline of the training loss at early epoch $e$ across several different independent runs. 

\textbf{Herding.} This method~\cite{welling2009herding,chen2012super} selects samples based on the distance between the coreset center and original dataset center in the feature space. It incrementally adds $1$ sample each time into the coreset that can minimize distance between two centers.

In this paper, we reproduce these baselines following the instructions of Deepcore~\cite{guo2022deepcore} and compare these with our DUSE.


\begin{table*}[htbp]
  \centering
  \caption{Comparison of DUSE against coreset selection baselines under various expansion rates in the setting of class imbalance. The best results are marked in \textbf{boldface}, and the sub-optimal ones are \underline{underlined}. For each experimental result, we run $3$ times and report its mean and standard deviation.}
  \resizebox{0.9\textwidth}{!}{
    \begin{tabular}{ccccccc}
    \toprule
    \multirow{2}[4]{*}{Dataset} & \multirow{2}[4]{*}{Methods} & \multicolumn{5}{c}{Expansion Rate} \\
\cmidrule{3-7}          &       & 1\%   & 4\%   & 7\%   & 9\%   & 19\% \\
    \midrule
    \multirow{9}[2]{*}{RML2016.10a} & Least confidence & \underline{49.29$\pm$1.32} & 50.55$\pm$1.92 & 49.72$\pm$0.44 & 46.26$\pm$4.13 & 53.09$\pm$2.39 \\
          & Entropy & 46.98$\pm$0.15 & 50.26$\pm$2.69 & 51.48$\pm$5.08 & 50.59$\pm$3.17 & 48.85$\pm$0.26 \\
          & Margin & 43.61$\pm$4.93 & 47.76$\pm$7.51 & 48.77$\pm$2.18 & 49.56$\pm$3.80 & 48.31$\pm$0.80 \\
          & DeepFool & 46.77$\pm$0.39 & 52.96$\pm$2.27 & 48.93$\pm$5.09 & 52.22$\pm$2.61 & 54.10$\pm$0.87 \\
          & Forgetting & \textbf{51.80$\pm$2.42} & \underline{62.93$\pm$3.93} & \underline{66.87$\pm$4.03} & \underline{69.93$\pm$2.16} & \underline{73.65$\pm$1.95} \\
          & Glister & 45.52$\pm$1.73 & 46.93$\pm$6.08 & 48.46$\pm$1.39 & 44.73$\pm$2.08 & 47.48$\pm$3.43 \\
          & GraNd & 38.99$\pm$2.40 & 44.84$\pm$9.11 & 45.94$\pm$4.34 & 43.17$\pm$9.68 & 49.87$\pm$5.94 \\
          & Herding & 37.34$\pm$1.59 & 39.45$\pm$2.41 & 41.31$\pm$1.88 & 42.99$\pm$1.37 & 47.22$\pm$1.98 \\
          & \cellcolor[rgb]{ .839,  .863,  .894}DUSE & \cellcolor[rgb]{ .839,  .863,  .894}48.68$\pm$1.44 & \cellcolor[rgb]{ .839,  .863,  .894}\textbf{68.07$\pm$2.41} & \cellcolor[rgb]{ .839,  .863,  .894}\textbf{72.26$\pm$1.01} & \cellcolor[rgb]{ .839,  .863,  .894}\textbf{75.18$\pm$2.11} & \cellcolor[rgb]{ .839,  .863,  .894}\textbf{80.69$\pm$0.18} \\
    \midrule
    \multirow{9}[2]{*}{Sig2019-12} & Least confidence & 23.67$\pm$0.16 & 38.80$\pm$9.92 & 48.05$\pm$3.28 & 46.50$\pm$2.28 & 35.13$\pm$19.65 \\
          & Entropy & 18.94$\pm$7.53 & 18.83$\pm$7.42 & 32.84$\pm$17.37 & 32.11$\pm$16.86 & 46.50$\pm$2.56 \\
          & Margin & \underline{30.69$\pm$9.43} & 47.61$\pm$2.47 & 34.32$\pm$18.53 & 42.68$\pm$1.63 & 47.09$\pm$3.74 \\
          & DeepFool & 18.92$\pm$0.76 & 8.33$\pm$0.00 & 11.50$\pm$4.48 & 17.80$\pm$13.39 & 19.42$\pm$10.29 \\
          & Forgetting & 21.62$\pm$1.57 & \textbf{49.20$\pm$1.32} & \underline{51.66$\pm$0.82} & \underline{55.33$\pm$0.45} & \underline{60.73$\pm$1.61} \\
          & Glister & 24.43$\pm$13.78 & 24.6$\pm$15.62 & 11.28$\pm$4.18 & 40.35$\pm$17.02 & 36.13$\pm$8.51 \\
          & GraNd & 17.77$\pm$6.82 & 21.78$\pm$2.02 & 12.39$\pm$5.73 & 38.52$\pm$12.31 & 29.51$\pm$15.20 \\
          & Herding & \textbf{38.33$\pm$12.0} & 33.84$\pm$12.51 & 8.33$\pm$0.00 & 19.75$\pm$16.15 & 33.23$\pm$18.17 \\
          & \cellcolor[rgb]{ .839,  .863,  .894}DUSE & \cellcolor[rgb]{ .839,  .863,  .894}20.98$\pm$0.27 & \cellcolor[rgb]{ .839,  .863,  .894}\underline{48.69$\pm$1.79} & \cellcolor[rgb]{ .839,  .863,  .894}\textbf{54.73$\pm$1.88} & \cellcolor[rgb]{ .839,  .863,  .894}\textbf{55.35$\pm$2.21} & \cellcolor[rgb]{ .839,  .863,  .894}\textbf{61.41$\pm$0.15} \\
    \midrule
    \multirow{9}[2]{*}{RML2018.10a} & Least confidence & 40.04$\pm$0.51 & 27.76$\pm$4.95 & 36.87$\pm$1.85 & 33.46$\pm$1.90 & 39.26$\pm$1.47 \\
          & Entropy & 30.94$\pm$12.92 & 37.30$\pm$4.20 & 33.02$\pm$8.07 & 28.58$\pm$4.99 & 40.41$\pm$3.45 \\
          & Margin & 31.61$\pm$13.43 & 22.99$\pm$9.10 & 25.75$\pm$4.02 & 36.89$\pm$3.51 & 42.38$\pm$0.46 \\
          & DeepFool & 25.51$\pm$3.13 & 10.72$\pm$4.64 & 33.87$\pm$0.86 & 34.13$\pm$0.53 & 38.07$\pm$3.85 \\
          & Forgetting & \underline{44.81$\pm$1.83} & \underline{48.59$\pm$1.22} & \underline{49.25$\pm$1.68} & \underline{48.32$\pm$1.72} & \textbf{55.86$\pm$0.38} \\
          & Glister & 33.57$\pm$10.29 & 17.25$\pm$6.93 & 36.60$\pm$0.97 & 39.29$\pm$1.96 & 42.44$\pm$2.14 \\
          & GraNd & 36.16$\pm$3.06 & 40.51$\pm$2.29 & 34.06$\pm$5.01 & 40.48$\pm$1.69 & 41.28$\pm$2.77 \\
          & Herding & 36.85$\pm$3.76 & 37.67$\pm$3.39 & 20.85$\pm$12.63 & 39.01$\pm$4.58 & 43.69$\pm$2.40 \\
          & \cellcolor[rgb]{ .839,  .863,  .894}DUSE & \cellcolor[rgb]{ .839,  .863,  .894}\textbf{47.67$\pm$0.59} & \cellcolor[rgb]{ .839,  .863,  .894}\textbf{49.12$\pm$0.93} & \cellcolor[rgb]{ .839,  .863,  .894}\textbf{49.56$\pm$1.72} & \cellcolor[rgb]{ .839,  .863,  .894}\textbf{49.61$\pm$1.21} & \cellcolor[rgb]{ .839,  .863,  .894}\underline{50.45$\pm$0.17} \\
    \bottomrule
    \end{tabular}}
  \label{tab:mainexp}%
\end{table*}%

\begin{table*}[t]
  \centering
 \caption{Comparison of DUSE against coreset selection baselines under various expansion rates in the setting of class balance. The best results are marked in \textbf{boldface}, and the sub-optimal ones are \underline{underlined}. For each experimental result, we run $3$ times and report its mean and standard deviation.}
  \resizebox{0.9\textwidth}{!}{
    \begin{tabular}{ccccccc}
    \toprule
    \multirow{2}[3]{*}{Dataset} & \multirow{2}[3]{*}{Methods} & \multicolumn{5}{c}{Expansion Rate} \\
\cmidrule{3-7}          &       & 1\%   & 4\%   & 7\%   & 9\%   & 19\% \\
    \midrule
    \multirow{9}[2]{*}{RML2016.10a} & Least confidence & 53.89$\pm$0.74 & 65.39$\pm$1.50 & 70.00$\pm$0.35 & 71.45$\pm$0.90 & 76.29$\pm$0.05 \\
          & Entropy & \underline{54.56$\pm$0.65} & 65.51$\pm$0.51 & \underline{70.46$\pm$0.29} & 71.09$\pm$0.63 & 75.68$\pm$0.48 \\
          & Margin & 54.10$\pm$0.15 & 65.17$\pm$1.30 & 70.38$\pm$0.34 & 70.88$\pm$0.77 & 76.10$\pm$0.71 \\
          & DeepFool & \textbf{54.61$\pm$0.75} & 65.11$\pm$2.46 & 70.01$\pm$0.08 & 70.97$\pm$0.58 & \underline{76.49$\pm$0.25} \\
          & Forgetting & 52.66$\pm$0.93 & 65.79$\pm$0.56 & 70.24$\pm$0.34 & \underline{72.10$\pm$0.51} & 75.99$\pm$0.53 \\
          & Glister & 52.26$\pm$0.48 & 64.90$\pm$0.48 & 67.66$\pm$1.77 & 70.88$\pm$0.38 & 76.39$\pm$0.55 \\
          & GraNd & 52.47$\pm$0.60 & 60.02$\pm$5.37 & 63.70$\pm$4.11 & 69.27$\pm$3.86 & 74.63$\pm$1.55 \\
          & Herding & 50.44$\pm$1.48 & \underline{66.01$\pm$0.74} & 69.08$\pm$0.58 & 70.86$\pm$0.33 & 76.31$\pm$0.96 \\
          & \cellcolor[rgb]{ .839,  .863,  .894}DUSE & \cellcolor[rgb]{ .851,  .851,  .851}52.15$\pm$0.48 & \cellcolor[rgb]{ .851,  .851,  .851}\textbf{66.53$\pm$0.75} & \cellcolor[rgb]{ .851,  .851,  .851}\textbf{71.56$\pm$0.26} & \cellcolor[rgb]{ .851,  .851,  .851}\textbf{73.36$\pm$0.13} & \cellcolor[rgb]{ .851,  .851,  .851}\textbf{77.06$\pm$0.27} \\
    \bottomrule
    \end{tabular}}
  \label{tab:mainexp balance}%
\end{table*}%

\begin{table*}[htbp]
  \centering
  \caption{Cross-architecture generalization of the expanded dataset obtained by DUSE. The expanded dataset is obtained using 2D-CNN at an expansion rate of $4\%$ in the setting of class imbalance and is verified on different model structures. For each experimental result, we run $3$ times and report its mean and standard deviation.}
  \resizebox{0.99\textwidth}{!}{
    \begin{tabular}{ccccccc}
    \toprule
    \multirow{2}[3]{*}{Dataset} & \multicolumn{6}{c}{Evaluation Model } \\
\cmidrule{2-7}          & 1D-CNN & 2D-CNN & AlexNet & SigNet & GRU   & MCLDNN \\
    \midrule
    RML2016.10a & 67.35$\pm$0.74 & 68.07$\pm$2.41 & 81.01$\pm$0.50 & 80.64$\pm$1.92 & 71.49$\pm$0.25 & 74.90$\pm$0.94 \\
    Sig2019-12 & 73.00$\pm$0.43 & 48.69$\pm$1.79 & 77.78$\pm$3.43 & 93.10$\pm$0.38 & 43.77$\pm$0.69 & 48.88$\pm$2.01 \\
    RML2018.10a & 64.21$\pm$0.88 & 49.12$\pm$0.93 & 64.98$\pm$0.12 &    81.71$\pm$0.70   & 44.51$\pm$0.25 & 69.45$\pm$2.60 \\
    \bottomrule
    \end{tabular}}
  \label{tab:cross model}%
\end{table*}%

\subsection{Models} 
In this paper, we use 2D‑CNN to perform uncertainty scoring and evaluate the expanded dataset $\mathcal{D}_{T}^*$ on 2D‑CNN by default. As for the cross-architecture generalization experiments, we utilize 1D-CNN~\cite{o2018over}, SigNet~\cite{chen2021signet}, AlexNet~\cite{krizhevsky2012imagenet}, GRU, MCLDNN~\cite{xu2020spatiotemporal} to conduct evaluation.

\subsection{Implementation details}
For a fair comparison, we set the same hyperparameter settings to train models with the expanded dataset obtained by baseline methods and our DUSE. Specifically, as for evaluation, we use a learning rate of $0.001$ and a batch size of $128$ for $50$ epochs. In the data expansion stage, we also use a learning rate of $0.001$ and a batch size of $128$, but only train for $20$ epochs. For each experimental result, we run $3$ times and report its mean and standard deviation to avoid the randomness. All experiments are conducted on two NVIDIA Tesla A100 GPUs.

\subsection{Main Experiments}

\begin{figure*}[t]
	\centering
    \includegraphics[width=0.99\textwidth]{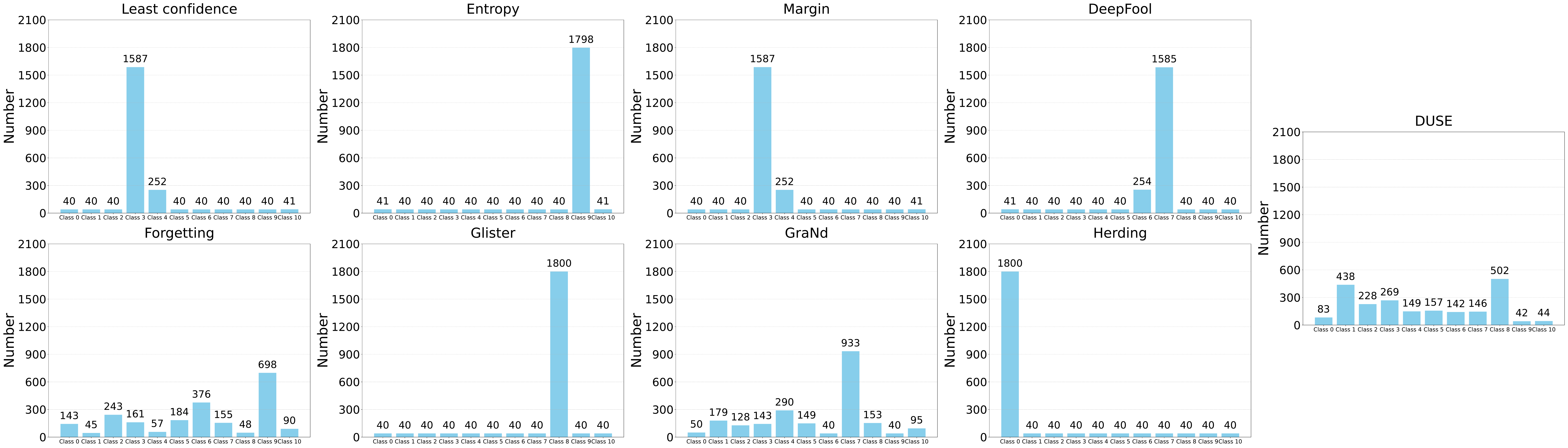} 
 \caption{Visualization of the number of samples per class in the expanded dataset obtained by various coreset selection baselines and DUSE (under the setting of class imbalance). Experiments are conducted on RML2016.10a under the $4\%$ expansion rate.} 
 \label{fig:class number}
\end{figure*}

\textbf{Effectiveness of DUSE under the setting of class imbalance.} To evaluate the effectiveness of DUSE, we compare it with $8$ coreset selection baselines  (Least Confidence, Entropy, Margin, DeepFool, Forgetting, Glister, GraNd, Herding) on $3$ standard AMR benchmarks: RML2016.10a, Sig2019‑12, and RML2018.10a. For each dataset, we vary the expansion rate $r$ over $\{1 \%, 4 \%, 7 \%, 9 \%, 19 \%\}$, where $r$ can be formulated as follows:
\begin{equation}
r = \frac{|\mathcal{D}_{T}^*-\mathcal{D}_{T}|}{\mathcal{D}_A} 
\end{equation}
Besides, as mentioned in \cref{sec:uncertainty}, DUSE needs a deep learning model to quantify how “informative” each sample in $\mathcal{D}_A$ is. Therefore, in this paper, we uniformly use 2D‑CNN to perform uncertainty scoring and evaluate the expanded dataset $\mathcal{D}_{T}^*$ on 2D‑CNN. We report the mean and standard deviation over $3$ random trials. Under the setting of class imbalance, DUSE only selects a fixed number of samples based on the uncertainty scoring function, without any class‐balance constraints. In other words, the selected samples may have very different numbers of each category (see \cref{fig:class number} for details). As shown in \cref{tab:mainexp}, Across all three datasets and most expansion rates, DUSE significantly outperforms the existing coreset selection baselines. Specifically, on the RML2016.10a dataset, when the expansion rate is $1\%$, the accuracy of DUSE is $48.68\pm1.44$, while the best forgetting baseline reaches $51.80\pm2.42$, $3.12\%$ higher than our method. When the expansion rate is increased to $4\%$, DUSE jumps to $68.07\pm2.41$, $5.06\%$ higher than forgetting’s $62.93\pm3.93$; when the expansion rate is further expanded to $7\%$, the gap between the two remains at $5.39\%$ ($72.26\pm1.01$ vs. $66.87\pm4.03$). When the expansion rate reaches $9\%$, DUSE leads forgetting with $75.18\pm2.11$ and $69.93\pm2.16$ respectively; and at the highest expansion rate of $19\%$, DUSE's $80.69\pm0.18$ is $7.04\%$ higher than forgetting's $73.65\pm1.95$, showing stronger scalability. Turning to the Sig2019‑12 dataset, with only $1\%$ auxiliary samples, DUSE achieves $20.98\pm0.27$, lower than Herding's $38.33\pm12.0$ and margin's $30.69\pm9.43$. When the expansion rate is increased to $4\%$, the two are almost on par. DUSE is $48.69\pm1.79$, while forgetting is $49.20\pm1.32$. When it continued to increase to $7\%$, DUSE's accuracy once again surpasses its opponent, reaching $54.73\pm1.88$ (forgetting is $51.66\pm0.82$), leading by $3.07\%$. Further expanding to $9\%$, the performance of the two methods is almost the same ($55.35\pm2.21$ vs. $55.33\pm0.45$); at the highest $19\%$ budget, DUSE leads slightly with $61.41\pm0.15$ beating forgetting's $60.73\pm1.61$. As for the RML2018.10a dataset, in the low-budget scenario ($1\%$ expansion rate), DUSE surpasses forgetting's $44.81\pm1.83$ with $47.67\pm0.59$, leading by nearly $2.9\%$; when the expansion rate is $4\%$, DUSE is slightly better than forgetting ($49.12\pm0.93$ vs. $48.59\pm1.22$). When the expansion rate reaches $7\%$ and $9\%$, DUSE is slightly better than forgetting ($49.25\pm1.68$, $48.32\pm1.72$) with $49.56\pm1.72$ and $49.61\pm1.21$, respectively. At the maximum budget of $19\%$, forgetting surpasses DUSE by nearly $5.4\%$. In general, although DUSE does not reach the optimal result under certain expansion rates, it achieves the best effect in most cases, which demonstrates the effectiveness of the proposed method under the setting of class imbalance. In \cref{sec:Additional Experiments and Analyses}, we further delve into the reasons why our method achieves good results compared to other coreset selection baselines under the setting of class imbalance.

\textbf{Effectiveness of DUSE under the setting of class balance.} In the previous paragraph, we have demonstrated the effectiveness of DUSE under the setting of class imbalance. Herein, we further explore the effectiveness of DUSE under the setting of class balance. We follow the same experimental setup as in the previous section, except that we ensure class balance (i.e., ensure that the number of samples in each category of the expanded dataset is the same). As shown in \cref{tab:mainexp balance}, DUSE consistently outperforms existing coreset selection baselines, which demonstrates the effectiveness of DUSE under the setting of class balance.

\textbf{Cross-architecture Generalization.}
As mentioned in \cref{sec:uncertainty}, DUSE needs a deep learning model to quantify how “informative” each sample in $\mathcal{D}_A$ is. In this paper, we uniformly use 2D‑CNN to perform uncertainty scoring and evaluate the expanded dataset $\mathcal{D}_{T}^*$ on 2D‑CNN. Herein, to validate that our selection strategy is not tied to a specific backbone, we further assess its cross-architecture generalization across $5$ additional architectures (1D-CNN~\cite{o2018over}, SigNet~\cite{chen2021signet}, AlexNet~\cite{krizhevsky2012imagenet}, GRU, MCLDNN~\cite{xu2020spatiotemporal}) on $3$ AMR datasets (RML2016.10a, Sig2019-12, RML2018.10a). Specifically, we use the expanded dataset $\mathcal{D}_{T}^*$ obtained by 2D‑CNN to train different models and report their accuracy on the test dataset. As shown in \cref{tab:cross model}, despite being selected for a specific architecture (2D‑CNN), our expanded dataset does not seem to suffer from much over-fitting to that model. Notably, on the Sig2019-12 dataset, SigNet achieves an accuracy of $93.10\%$ using our expanded dataset (only $5\%$ of the original dataset).

\begin{figure*}[htbp]
	\centering
    \includegraphics[width=0.99\textwidth]{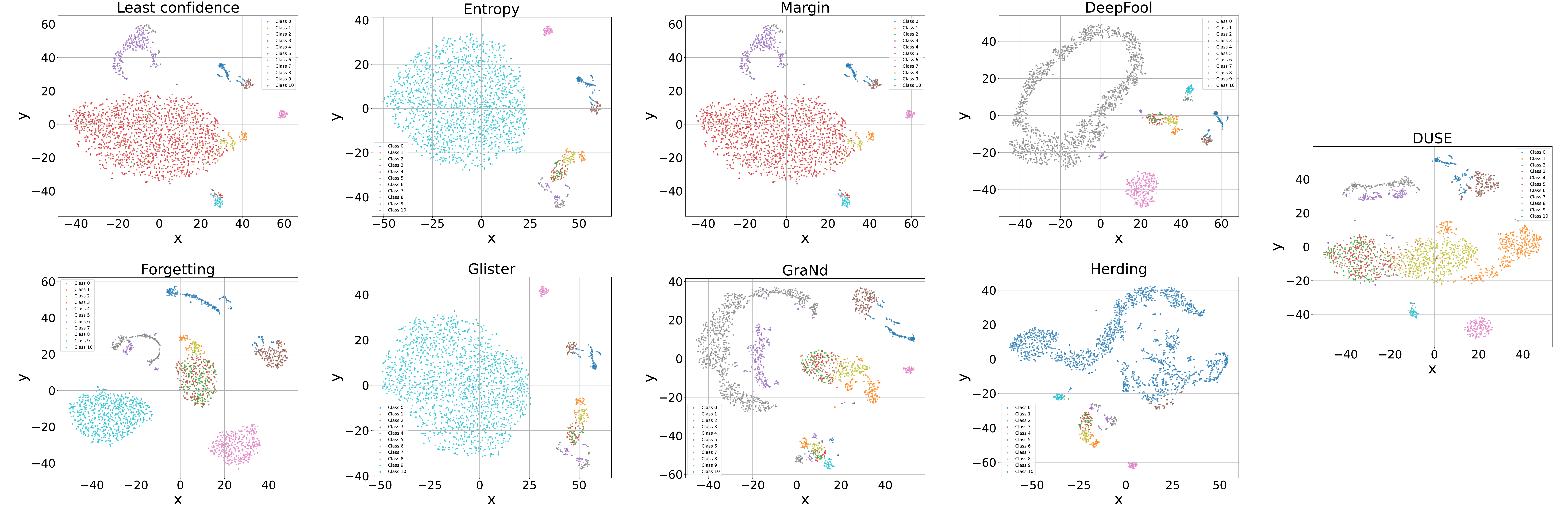} 
 \caption{T-SNE visualization of the expanded dataset obtained by various coreset selection baselines and DUSE (under the setting of class imbalance). Experiments are conducted on RML2016.10a under the $4\%$ expansion rate.} 
 \label{fig:tsne}
\end{figure*}

\subsection{Additional Experiments and Analyses}
\label{sec:Additional Experiments and Analyses}
To further analyse the effectiveness of our DUSE framework, we compare the number of samples of each category in $\mathcal{D}_{T}^*$ obtained by each method, as well as their corresponding t-Distributed Stochastic Neighbor Embedding~\cite{hinton2002stochastic} (t-SNE) visualizations. Specifically, we conduct experiments on RML2016.10a under the $4\%$ expansion rate.

\textbf{Visualization of the number of samples per class.}
Investigating the number of each class is crucial because uneven selection (some classes are heavily over-selected while others are rarely selected) can introduce bias to the augmented dataset and degrade recognition performance on under-sampled classes. Therefore, for each method, we first count the number of samples of each modulation category in $\mathcal{D}_{T}^*$ and present them as a grouped bar chart. As shown in \cref{fig:class number}, we observe that DUSE maintains the class balance of target dataset more faithfully than other methods, which usually focus on a few classes. This uniform class distribution demonstrates that DUSE not only recognizes more diverse samples, but also follows the inherent distribution of $\mathcal{D}_T$, thus achieving better performance.

\textbf{T-SNE visualization.} T‑SNE visualization allows us to intuitively evaluate the manifold distribution of the expanded dataset obtained by a given method. Therefore, we use a 2D-CNN trained on the whole training dataset to extract feature representations of the augmented dataset obtained by each method and visualize them using t‑SNE. As shown in \cref{fig:tsne}, DUSE has similar point density across all classes, and this balanced coverage avoids bias towards any single modulation type and enhances the model's generalization across all classes. This phenomenon can also be seen in \cref{fig:class number}. Besides, DUSE also ensures that samples of the same category form tight clusters, demonstrating the excellent category discrimination ability of the proposed method.

\subsection{Ablation Study}
\label{sec: Ablation Study}
In this subsection, we conduct detailed ablation experiments on the effect of active learning in DUSE under the setting of class imbalance. To quantify the benefit of incorporating an active learning loop into our DUSE framework, we compare two variants:  
\begin{itemize}  
  \item \textbf{DUSE (w/o active learning)}: select samples in one pass using uncertainty scoring function from a 2D‑CNN trained on $\mathcal{D}_T$.  
  \item \textbf{DUSE (w active learning)}: alternate between (1) training the 2D‑CNN on the currently augmented dataset and (2) re‑scoring and selecting  samples using the currently 2D‑CNN in the updated auxiliary dataset, until the total budget is reached.  
\end{itemize}
Specifically, we conduct experiments using these two variants on RML2016.10a, Sig2019-12, RML2018.10a under the $7\%$ expansion rate. As shown in \cref{tab:aba}, the use of active learning strategies significantly improves the quality of the extended dataset, leading to significantly improved accuracy on the three AMR datasets. We attribute this improvement to the iterative updates of the model on the progressively augmented dataset: by retraining on newly added samples, the model continually refines its uncertainty estimates to adapt to the evolving data distribution. This dynamic feedback mitigates early sorting errors and ensures that, at each stage, only the most informative samples are selected.

\begin{table}[t]
  \centering
  \caption{Ablation experiments on the effect of active learning in DUSE under the setting of class imbalance. Experiments are conducted using 2D-CNN under the $7\%$ expansion rate. For each experimental result, we run $3$ times and report its mean and standard deviation.}
    \begin{tabular}{c|ccc}
    \toprule
    \multirow{2}[4]{*}{Active Learning} & \multicolumn{3}{c}{Acc} \\
\cmidrule{2-4}          & \multicolumn{1}{c}{RML2016.10a} & \multicolumn{1}{c}{Sig2019-12} & \multicolumn{1}{c}{RML2018.10a} \\
    \midrule
    \XSolid      & \multicolumn{1}{c}{48.77$\pm$2.18} & \multicolumn{1}{c}{34.32$\pm$18.53} & \multicolumn{1}{c}{25.75$\pm$4.02} \\
    \Checkmark    & \multicolumn{1}{c}{72.26$\pm$1.01} & \multicolumn{1}{c}{54.73$\pm$1.88} & \multicolumn{1}{c}{49.56$\pm$1.72} \\
    Gain  &     $\pm$23.49  &   $\pm$20.41    & $\pm$23.81 \\
    \bottomrule
    \end{tabular}%
  \label{tab:aba}%
\end{table}%

\section{Conclusion}
\label{sec:Conclusion}
In this paper, we propose DUSE, a novel data expansion framework, aims to augment with samples drawn from large-scale AMR datasets to enlarge the target dataset. Specifically, we introduce an uncertainty scoring function to quantify the contribution of each sample in the auxiliary dataset and utilize an active learning strategy to iteratively update the deep learning model for scoring. By doing these, DUSE achieves superior performance compared to existing baselines in both class-balanced and class-imbalanced settings and has better cross-architecture generalization.

\bibliographystyle{IEEEtran}
\bibliography{reference}

\begin{IEEEbiography}[{\includegraphics[width=1in,height=1.25in,clip,keepaspectratio]{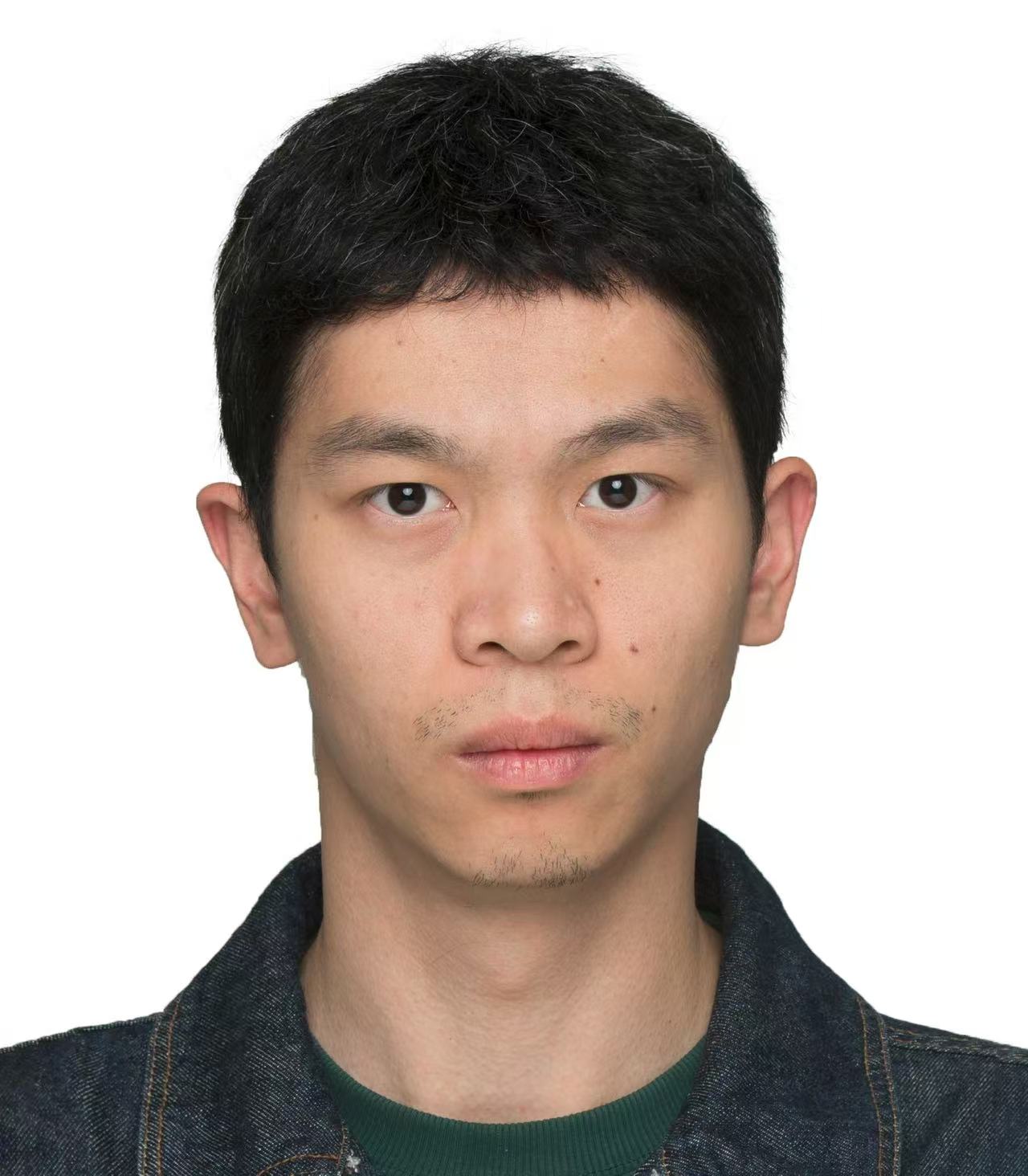}}]{Yao Lu}
received his B.S. degree from Zhejiang University of Technology and is currently pursuing a Ph.D. in control science and engineering at Zhejiang University of Technology. He is a visiting scholar with the Centre for Frontier AI Research, Agency for Science, Technology and Research, Singapore. He has published several academic papers in international conferences and journals, including ECCV, TNNLS, Neurocomputing and TCCN. He serves as a reviewer of ICLR2025, CVPR2025, ICCV25, NeurIPS25 and TNNLS. His research interests include deep learning and computer vision, with a focus on artificial intelligence and model compression.
\end{IEEEbiography} 
\vspace{-15mm}
\begin{IEEEbiography}[{\includegraphics[width=1in,height=1.25in,clip,keepaspectratio]{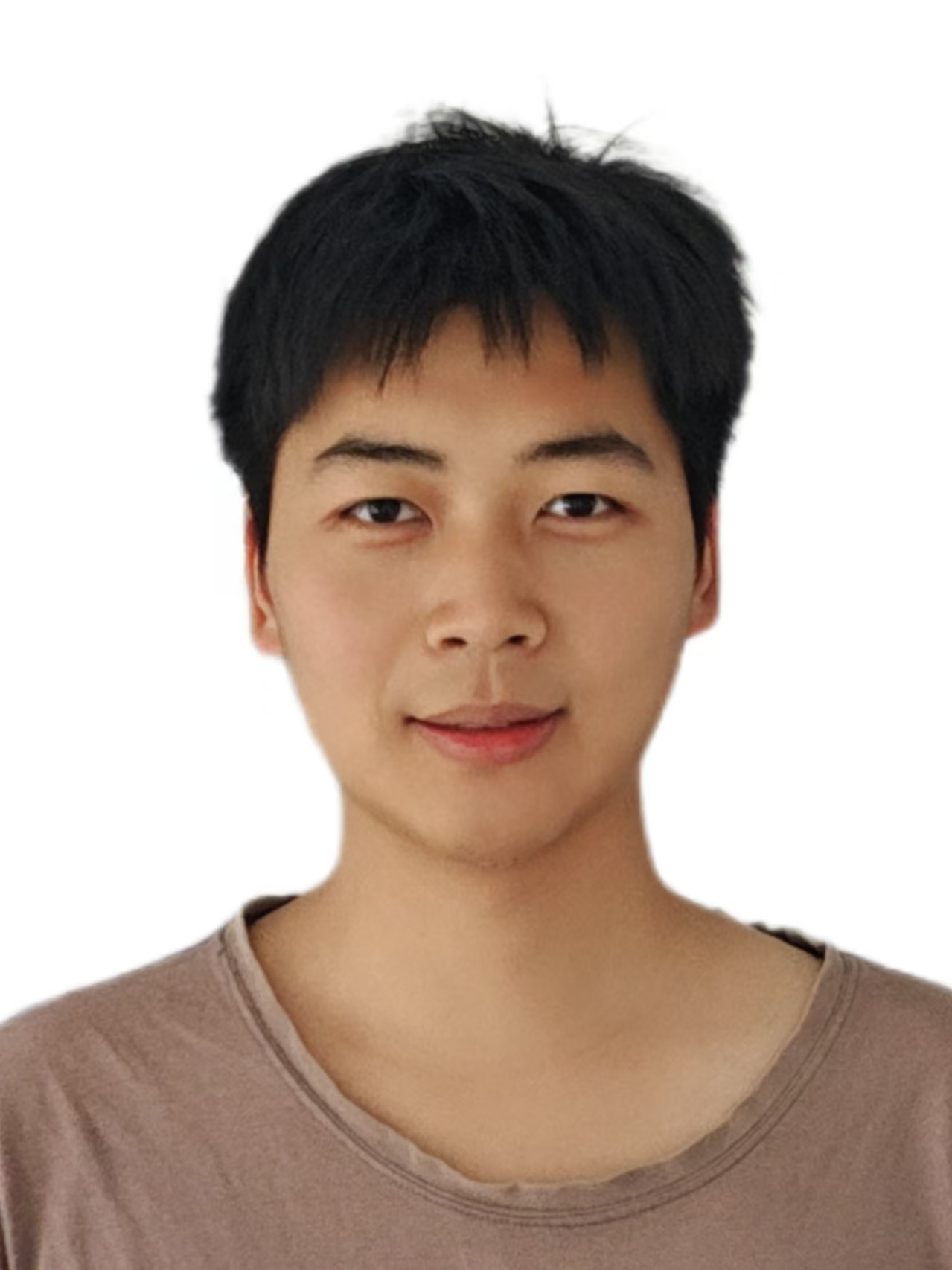}}]{Hongyu Gao}
is currently pursuing a Master's degree in Communication Engineering at Zhejiang University of Technology. His research interests include deep learning and data augmentation, with a focus on artificial intelligence and data compression.
\end{IEEEbiography}
\vspace{-15mm}

\begin{IEEEbiography}[{\includegraphics[width=1in,height=1.25in,clip,keepaspectratio]{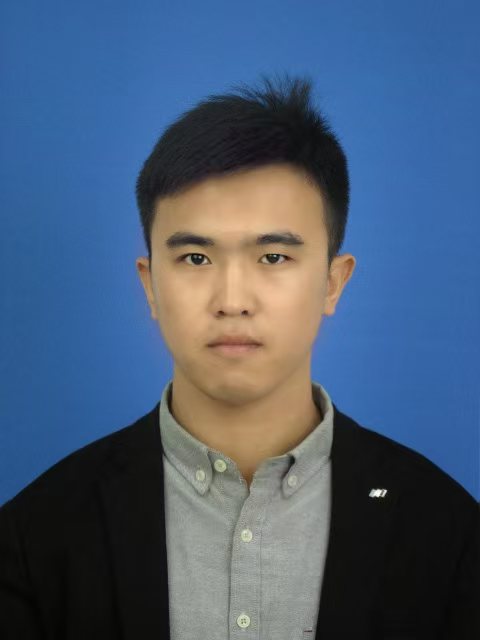}}]{Zhuangzhi Chen}
received the B.S. and Ph.D. degrees in Control Theory and Engineering from Zhejiang University of Technology, Hangzhou, China, in 2017 and 2022, respectively. He was a visiting scholar with the Department of Computer Science, University of California at Davis, CA, in 2019. He completed his postdoctoral research in Computer Science and Technology at Zhejiang University of Technology, in 2024. He is currently working as an associate researcher at the Cyberspace Security Research Institute of Zhejiang University of Technology, and also serving as the deputy director and associate researcher at the Research Center of Electromagnetic Space Security, Binjiang Institute of Artificial Intelligence, ZJUT, Hangzhou 310056, China.

His current research interests include deep learning algorithm, deep learning-based radio signal recognition, and security AI of wireless communication.
\end{IEEEbiography} 

\vspace{-15mm}
\begin{IEEEbiography}[{\includegraphics[width=1in,height=1.25in,clip,keepaspectratio]{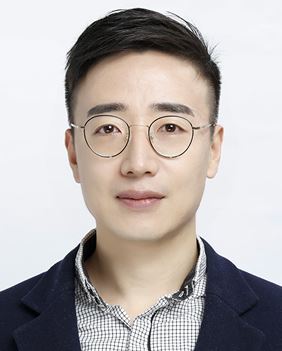}}]{Dongwei Xu}
(Member, IEEE) received the B.E. and Ph.D. degrees from the State Key Laboratory of Rail Traffic Control and Safety, Beijing Jiaotong University, Beijing, China, in 2008 and 2014, respectively. He is currently an Associate Professor with the Institute of Cyberspace Security, Zhejiang University of Technology, Hangzhou, China. His research interests include intelligent transportation Control, management, and traffic safety engineering.
\end{IEEEbiography}
\vspace{-15mm}
\begin{IEEEbiography}[{\includegraphics[width=1in,height=1.25in,clip,keepaspectratio]{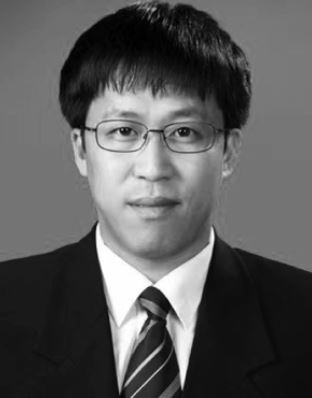}}]{Yun Lin}
(Member, IEEE) received the B.S. degree in electrical engineering from Dalian Maritime University, Dalian, China, in 2003, the M.S. degree in communication and information system from the Harbin Institute of Technology, Harbin, China, in 2005, and the Ph.D. degree in communication and information system from Harbin Engineering University, Harbin, in 2010. From 2014 to 2015, he was a Research Scholar with Wright State University, Dayton, OH, USA. He is currently a Full Professor with the College of Information and Communication Engineering, Harbin Engineering University. He has authored or coauthored more than 200 international peer-reviewed journal/conference papers, such as IEEE Transactions on Industrial Informatics, IEEE Transactions on Communications, IEEE Internet of Things Journal, IEEE Transactions on Vehicular Technology, IEEE Transactions on Cognitive Communications and Networking, TR, INFOCOM, GLOBECOM, ICC, VTC, and ICNC. His current research interests include machine learning and data analytics over wireless networks, signal processing and analysis, cognitive radio and software-defined radio, artificial intelligence, and pattern recognition.
\end{IEEEbiography}

\begin{IEEEbiography}[{\includegraphics[width=1in,height=1.25in,clip,keepaspectratio]{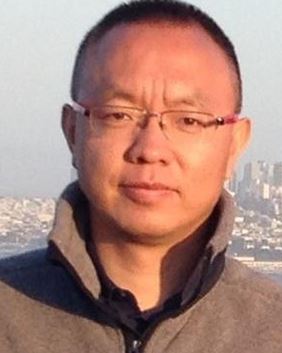}}]{Qi Xuan}
(Senior Member, IEEE) received the B.S. and Ph.D. degrees in control theory and engineering from Zhejiang University, Hangzhou, China, in 2003 and 2008, respectively. He was a Postdoctoral Researcher with the Department of Information Science and Electronic Engineering, Zhejiang University from 2008 to 2010, and a Research Assistant with the Department of Electronic Engineering, City University of Hong Kong, Hong Kong, in 2010 and 2017, respectively. From 2012 to 2014, he was a Postdoctoral Fellow with the Department of Computer Science, University of California at Davis, Davis, CA, USA. He is currently a Professor with the Institute of Cyberspace Security, College of Information Engineering, Zhejiang University of Technology, Hangzhou, and also with the PCL Research Center of Networks and Communications, Peng Cheng Laboratory, Shenzhen, China. He is also with Utron Technology Company Ltd., Xi’an, China, as a Hangzhou Qianjiang Distinguished Expert. His current research interests include network science, graph data mining, cyberspace security, machine learning, and computer vision.
\end{IEEEbiography}

\begin{IEEEbiography}[{\includegraphics[width=1in,height=1.25in,clip,keepaspectratio]{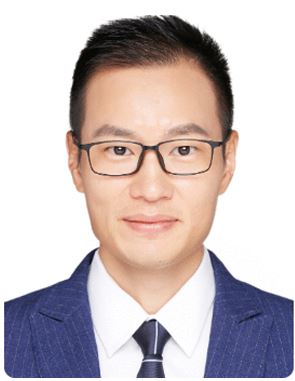}}]{Guan Gui}
    received the Dr. Eng. degree in information and communication engineering from the University of Electronic Science and Technology of China, Chengdu, China, in 2012. From 2009 to 2012, with financial support from the China Scholarship Council and the Global Center of Education, Tohoku University, he joined the Wireless Signal Processing and Network Laboratory (Prof. Adachis laboratory), Department of Communications Engineering, Graduate School of Engineering, Tohoku University, as a Research Assistant and a Post Doctoral Research Fellow, respectively. From 2012 to 2014, he was supported by the Japan Society for the Promotion of Science Fellowship as a Post Doctoral Research Fellow with the Wireless Signal Processing and Network Laboratory. From 2014 to 2015, he was an Assistant Professor with the Department of Electronics and Information System, Akita Prefectural University. Since 2015, he has been a Professor with the Nanjing University of Posts and Telecommunications, Nanjing, China. He is currently involved in the research of big data analysis, multidimensional system control, super-resolution radar imaging, adaptive filter, compressive sensing, sparse dictionary designing, channel estimation, and advanced wireless techniques. He received the IEEE International Conference on Communications Best Paper Award in 2014 and 2017 and the IEEE Vehicular Technology Conference (VTC-spring) Best Student Paper Award in 2014. He was also selected as a Jiangsu Special Appointed Professor, as a Jiangsu High-Level Innovation and Entrepreneurial Talent, and for 1311 Talent Plan in 2016. He has been an Associate Editor of the Wiley Journal Security and Communication Networks since 2012.
\end{IEEEbiography}



\vfill

\end{document}